\pdfoutput=1

\documentclass[11pt]{article}

\usepackage[]{acl}

\usepackage{times}

\usepackage{latexsym}
\usepackage{url}
\usepackage{latexsym}
\usepackage{multicol}
\usepackage{multirow}
  \usepackage{graphicx}  
  \usepackage{epstopdf}
  \usepackage{amssymb}
   \usepackage{color}
   \usepackage{amsmath} 
  \usepackage{xcolor}
  \usepackage{longtable}
\usepackage[T1]{fontenc}
\usepackage{hyperref}
\usepackage{xcolor}
\usepackage{times}
\usepackage{latexsym}
\usepackage{hyperref}[colorlinks,linkcolor=blue]
\usepackage[T1]{fontenc}
\usepackage{multirow}
\usepackage[utf8]{inputenc}
\usepackage{url}
\usepackage{microtype}
\usepackage{times}
\usepackage{helvet}
\usepackage{courier}
\usepackage{booktabs}
\usepackage{amssymb}
\usepackage[noend]{algpseudocode}   
\usepackage{hyperref}
\usepackage{algorithmicx,algorithm}
\usepackage[utf8]{inputenc}

\usepackage{microtype}

\title{Exploiting Sentiment and Common Sense for Zero-shot Stance Detection}
\author{
    Yun Luo \textsuperscript{\rm1,2},
    Zihan Liu \textsuperscript{\rm2},
    Yuefeng Shi \textsuperscript{\rm2},
    Stan Z. Li \textsuperscript{\rm2},
    Yue Zhang \textsuperscript{\rm2,3} 
    \\
        \textsuperscript{1} School of Computer Science And Technology, Zhejiang University, Hangzhou, 310024, P.R. China. \\
    \textsuperscript{2} School of Engineering, Westlake University, Hangzhou, 310024, P.R. China. \\
    \textsuperscript{3} Institute of Advanced Technology, Westlake Institute for Advanced Study, Hangzhou, 310024, P.R. China.  \\
    \texttt{\{luoyun, guofang, liuzihan, stan.zq.li, zhangyue\}@westlake.edu.cn}\\
}

\begin{document}
\maketitle
\begin{abstract}
The stance detection task aims to classify the stance toward given documents and topics. Since the topics can be implicit in documents and unseen in training data for zero-shot settings, we propose to boost the transferability of the stance detection model by using sentiment and commonsense knowledge, which are seldom considered in previous studies. Our model includes a graph autoencoder module to obtain commonsense knowledge and a stance detection module with sentiment and commonsense. Experimental results show that our model outperforms the state-of-the-art methods on the zero-shot and few-shot benchmark dataset--VAST. Meanwhile, ablation studies prove the significance of each module in our model. Analysis of the relations between sentiment, common sense, and stance indicates the effectiveness of sentiment and common sense.
\end{abstract}

\section{Introduction}
\begin{figure}[tpb]
      \centering
      \includegraphics[width=\hsize]{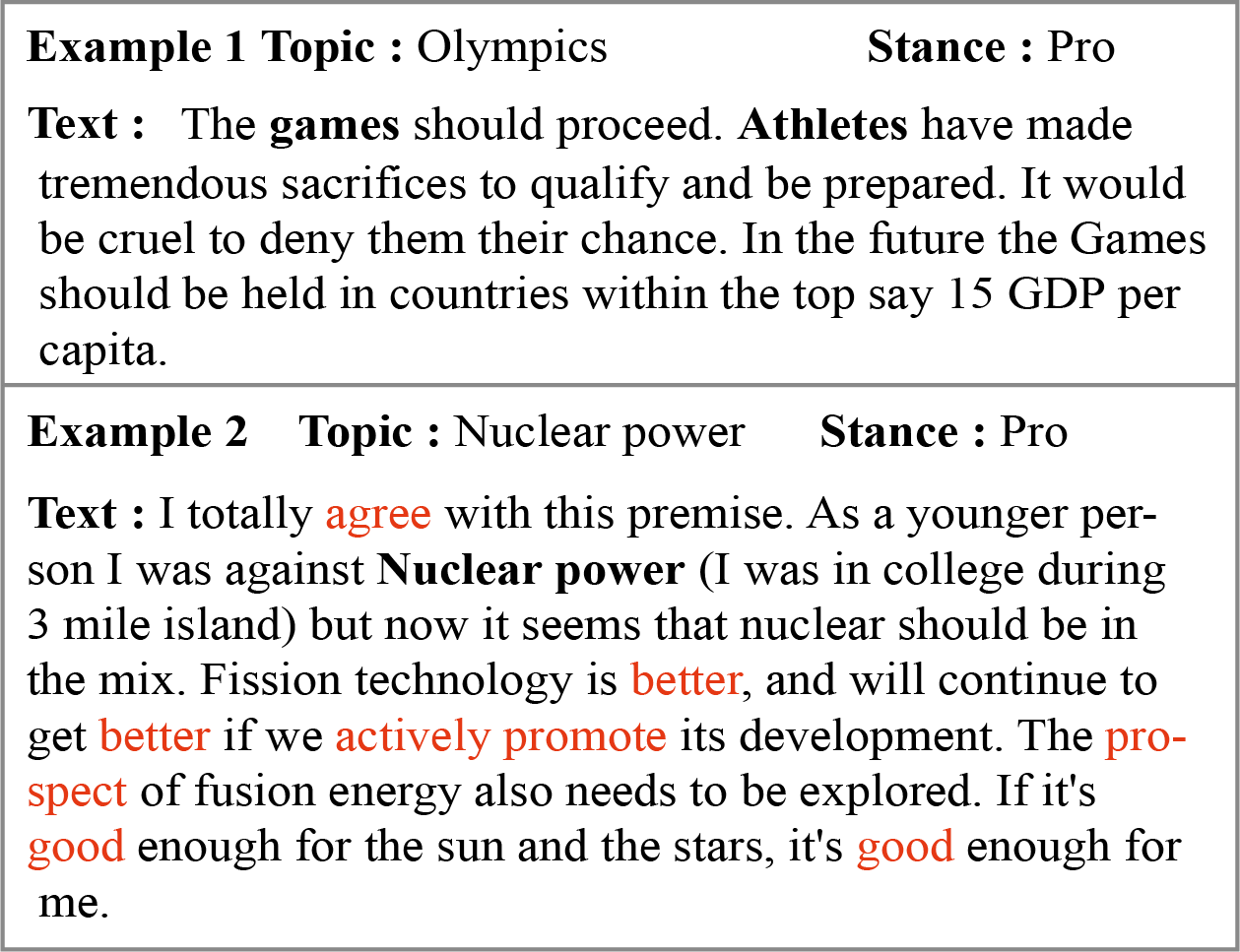}
      \caption{Examples for stance detection  VAST.}
      \label{fig1}
   \end{figure}

Stance detection aims to identify the authors' attitudes or positions (\textit{{Pro (support), Con (oppose), Neu (neutral)}}) towards a specific target such as an entity, a topic. \cite{Mohammad2017,mohammad-etal-2016-semeval,walker,qiu2015modeling,zhang2017we}. It is crucial for understanding opinions  and analyzing how opinions are presented in texts regarding specific issues, and
much work has been done building stance detection models  \cite{wei2016pkudblab,dias2016inf,allaway2020zero}.
There are two salient challenges to the task. First, obtaining rich annotated data in stance detection is time-consuming and labor-intensive. To address this issue, \citet{allaway2020zero} propose the dataset VAST containing various topics for few-shot and zero-shot stance detection tasks, requiring the model to classify the stance of topics unseen in the training set. Second, the topic is often not explicitly mentioned in the document, resulting in difficulty. Considering Figure \ref{fig1} Example 1, the document does not explicitly contain the topic `Olympics', but  `Games' and `Athlete' implicitly refer to the topic.

Existing work incorporates external knowledge to solve the challenges \cite{liu-etal-2021-enhancing,jayaram021human}. For example, CKE-Net achieves the state-of-the-art results for zero-shot stance detection, which uses pre-trained model BERT and commonsense knowledge graph on ConceptNet \cite{liu-etal-2021-enhancing}. However, such a method only considers the knowledge relations between documents and topics  (i.e., the commonsense knowledge in two-hop directed paths on the ConceptNet from documents to topics), 
limiting the generalization of adding other types of related knowledge. In Figure \ref{fig1} Example 1, the word `games' can also represent the computer programs in a different document. Such knowledge cannot be used for that document if no relation between `game' and `computer program' can be learned from the relations between documents and topics in the dataset.    

We consider incorporating two types of general knowledge, including common sense and sentiment. First, we incorporate commonsense knowledge into the stance detection model using a graph autoencoder module. We take a  pre-training method to train the graph autoencoder, separately to the stance detection module. 
Second, stance detection is significantly influenced by the sentiment information \cite{li-caragea-2019-multi,sobhani2016detecting,hardalov2022few} (case study can be seen in Appendix). In Figure 1 Example 2, the document contains many positive words like `good', and `better' regarding the topic `nuclear power', which implies a \textit{Pro} stance.
However, little existing work has considered sentiment knowledge for zero-shot stance detection. We use the sentiment-aware BERT (SentiBERT henceforth) to extract the sentiment information, assisting in classifying the stances of topics.




Existing work on injecting knowledge into NLP models can be broadly classified into two categories. One uses a graph encoder to integrate structural knowledge into a neural encoder \cite{li-etal-2019-improving,ghosal-etal-2020-kingdom,baietal} and the other injects knowledge by using training losses to tune model parameters \cite{jayaram021human,peters-etal-2019-knowledge,logan-etal-2019-baracks,liu-etal-2019-knowledge}. In our work, we consider the former for commonsense knowledge and the latter for sentiment due to the sources of information. In the component of knowledge graph encoding, a graph autoencoder  consisting of relational graph convolutional network (RGCN) encoders \cite{Schlichtkrull2018ModelingRD} and a DisMult decoder \cite{yang2014embedding} is trained using negative sampling to obtain the relations of concepts on the commonsense knowledge graph. We inject sentiment knowledge encoded by SentiBERT into BERT using a cross attention module and tuning the fusing process by the training loss of the stance detection.

      \begin{figure*}[tpb]
      \centering
      \includegraphics[width=0.93\hsize]{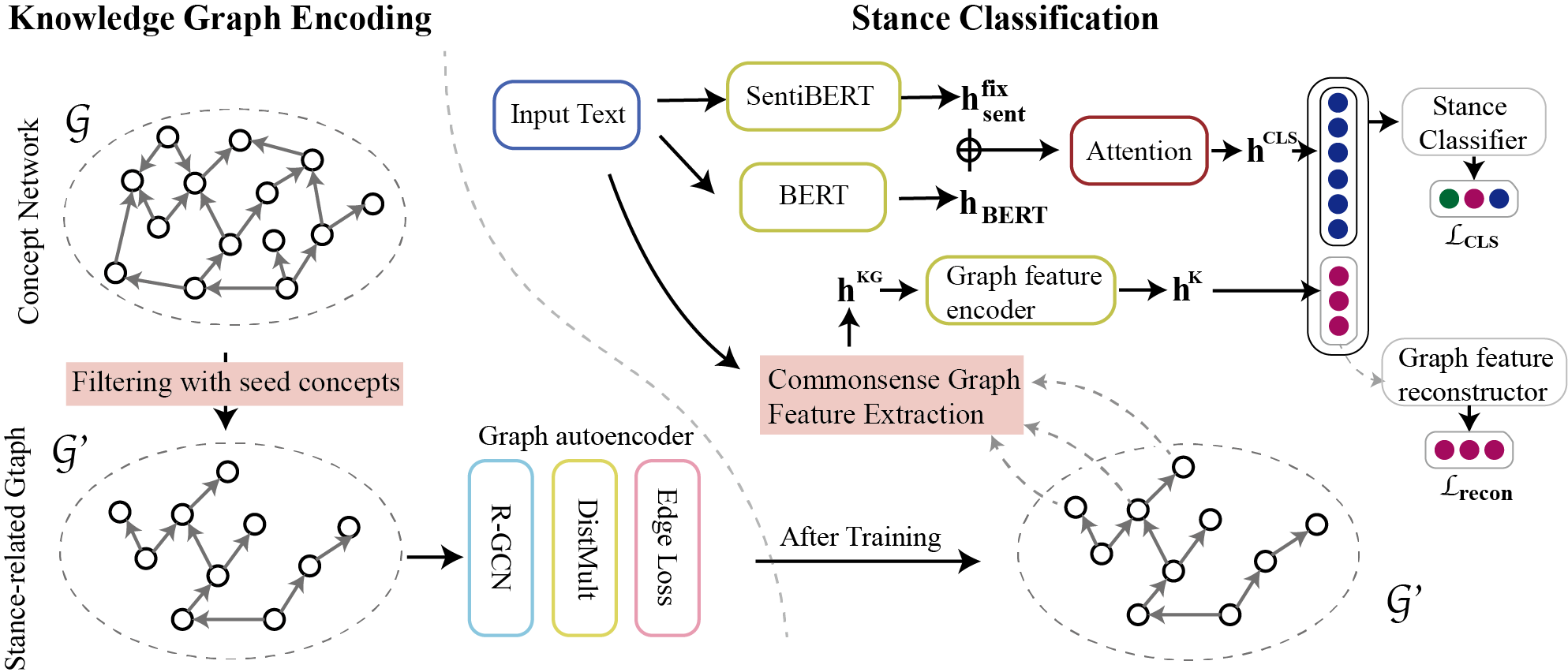}
      \caption{Framework of our proposed model, which contains two components, (1) knowledge graph encoding, (2) stance detection with sentiment and common sense.}
      \label{fig3}
   \end{figure*}
Our model achieves the state-of-the-art performance on the benchmark dataset VAST \cite{allaway2020zero}  in both zero-shot and few-shot stance detection, improving the performance on many challenging linguistic phenomena such as sarcasm and quotations. We analyze the performance of our model with respect to different sentiment and common sense features, finding that the data with the corresponding sentiment and stance pairs (i.e., \textit{(Pos, Pro)} and \textit{(Neg, Con)}) are the easiest part for models to classify; in addition, increased commonsense knowledge leads to improved performance of the stance detection model. To our knowledge, we are the first to incorporate both sentiment and common sense  into zero-shot stance detection model.
The code has been released \href{https://github.com/LuoXiaoHeics/StanceCS}{https://github.com/LuoXiaoHeics/StanceCS}.

\section{Related Work}
Stance detection, also known as stance classification \cite{walker}, stance identification \cite{zhang2017we}, stance prediction \cite{qiu2015modeling}, debate-side classification \cite{anand2011cats}, and debate stance classification \cite{hasan-ng-2013-stance}, aims to identify the stance of the text author towards a target (an entity, event, idea, opinion, claim, topic, etc.) either explicitly mentioned or implied within the text. For the initial task of stance detection, models are trained an individual classifier for each topic \cite{lin-etal-2006-side,beigman-klebanov-etal-2010-vocabulary,sridhar-etal-2015-joint,hasan-ng-2013-stance,hasan-ng-2014-taking,li-etal-2018-structured} or only a small number of topics are both in training and evaluation sets \cite{Faulkner2014AutomatedCO,du2017stance,hardalov-etal-2021-cross}.

However, given rich and varying topics, data annotation can be time-consuming and labor-intensive. Researchers attempt to solve the task in a cross-target setting \cite{augenstein-etal-2016-stance,xu2018cross}, training the model in a topic and testing it on another one, and propose several weakly supervised approaches using unlabeled data related to the test topics \cite{zarrella2016mitre,wei2016pkudblab,dias2016inf}.   Other studies propose the tasks of zero-shot and few-shot stance detection, which requires training the model in data of several topics and testing it on some unseen topics \cite{allaway2020zero}.

\citet{allaway2020zero}  propose to solve the task using a topic-grouped attention net, which uses the relation between the training and evaluation topics in an unsupervised way, and they also analyze the relationship between sentiment and stance from the perspective of the model by corrupting sentences with replacing sentiment words. \citet{jayaram021human} use human rationales as attribution priors to provide faithful explanations of models. \citet{liu-etal-2021-enhancing} propose to incorporate commonsense knowledge to learn the relations between different topics utilizing a CompGCN (a variant of graph convolution networks). However, it limits the content of knowledge (only knowledge from documents to stances in the training data). Our model differs from such a method in that our model adopts the related concepts of both documents and topics and uses a pre-trained graph autoencoder to obtain commonsense information.   Adversarial learning is also applied to solve the zero-shot task by using unlabeled raw data \cite{allaway2021adversarial}. Unlike the above work, we consider integrating external knowledge for zero-shot stance detection, including sentiment and commonsense information that are rarely considered. To our knowledge, we are the first to systematically incorporate sentiment and commonsense knowledge into the stance detection model and analyze the relationship between them (in Section 4.5 and 4.6).

\section{Method}
The architecture of our model is illustrated in Figure \ref{fig3}, which contains two components: (1) knowledge graph encoding, which integrates commonsense knowledge from ConceptNet (Section 3.1); (2) stance detection with sentiment and commonsense knowledge (Section 3.2).

\subsection{Knowledge Graph Autoencoder}
Formally, the ConceptNet is represented as a directed labeled graph $\mathcal{G = \{V,E,R}\}$,  with concepts $v_i \in \mathcal{V}$ and labeled edges $(v_i,r,v_j) \in \mathcal{E}$, where $r\in \mathcal{R}$ is the relation type of edge between $v_i$ and $v_j$. The concepts in ConceptNet are unigram words or n-gram phrases in the triplet format. For 
example, one such triplet from ConceptNet is \textit{(teacher, RelatedTo, job)}. 

ConceptNet has a large size of approximately 14 million edges. We extract a subset of edges related to the VAST dataset for our task. From the training documents in VAST, we first extract the set of all unique nouns, adjectives, and adverbs. These words are treated as the seeds that we use to filter the ConceptNet to a sub-graph. 
We extract all the triplets with a one-edge distance to any of those seed concepts, resulting in a sub-graph $\mathcal{G' = \{V',E',R'}\}$ with 310k concepts and 750k edges. The top 5 relations include `RelatedTo', `HasContext', `IsA', `Synonym' and `DerivedFrom'. The sub-graph $\mathcal{G'}$ contains all  the concepts related to stance targets in the VAST dataset. 

Following \citet{Schlichtkrull2018ModelingRD}, we construct a graph autoencoder to compute the representations of concepts in the sub-graph $\mathcal{G}'$. The autoencoder takes an incomplete set (randomly sampled with 50\% probability in our model) of edges $\hat{\mathcal{E}}'$ from $\mathcal{E}'$ in $\mathcal{G}'$ as input. $\hat{\mathcal{E}}'$ is negative sampled to the overall set of samples denoted $\mathcal{T}$ (details in Training). Then we assign the possible edges $(v_i,r,v_j) \in \mathcal{T}$  with scores to determine the probability these edges  are in $\mathcal{E}'$. Our graph autoencoder consists of a relational concept network (RGCN) \cite{Schlichtkrull2018ModelingRD} encoder to obtain the latent feature representations of concepts and a DistMult scoring decoder \cite{yang2014embedding} to recover the missing facts of triplets.

\textbf{Encoder}.
 RGCN  has a solid ability to accumulate relational evidence in multiple inference steps. In each step, a neighborhood-based convolutional feature transformation process uses the related concepts to induce an enriched stance-aggregated feature vector for each concept. Our model contains two stacked RGCN encoders. We first initialize the parameters of concept feature vectors $\textbf{g}_i$. Then the vectors are transformed into stance-aggregated feature vectors $\textbf{h}_i \in \mathbb{R}^d$ using the RGCN encoders:
\begin{equation}
    {f}(x_i,l) = \sigma(\sum_{r\in\mathcal{R}}\sum_{j\in N^r_i} \frac{1}{v_{i,r}}W_r^{(l)}x_j+W_0^{(l)}x_i),
    \nonumber 
\end{equation}

\begin{equation}
    \textbf{h}_i = \textbf{h}_i^{(2)} = f(\textbf{h}_i^{(1)},2) \ ; \ \textbf{h}_i^{(1)} = f(\textbf{g}_i,1),
\end{equation}
where $f$ is the encoder network (requiring inputs of feature vector $x_i$ and the rank of the layer $l$), $N^r_i$ denotes the neighbouring concepts $i$ with the relation $r\in \mathcal{R}$; $v_{i,r}$ is a normalization constant, which can be set in advance $v_{i,r} = |N^r_i|$ or learned by network learning; $\sigma$ is the activation function like ReLU and $W_r^{(1/2)},W_0^{(1/2)}$ are learnable parameters though training.

\textbf{Training}.  We use DistMult factorization as the decoder to assign scores. For a given triplet $(v_i,r,v_j) $, the score can be obtain as follows:
\begin{equation}
    s(v_i,r,v_j) = \sigma(\textbf{h}^T_{v_i}R_r\textbf{h}_{v_j}),
\end{equation}
where $\sigma$ is logistic function; $\textbf{h}_{v_i}, \textbf{h}_{v_j} \in \mathbb{R}^d$ are the encoding feature vectors through the graph encoder for concept $v_i$ and $v_j$. Each relation $r\in R$ is also associated with a diagonal matrix $R_r \in \mathbb{R}^{d\times d}$.

Our graph autoencoder module is trained using negative sampling \cite{Schlichtkrull2018ModelingRD}. We randomly corrupt the positive triplets, i.e., triplets in $\hat{\mathcal{E}}'$, to create an equal number of negative samples. The corruption is performed by modifying either of the connected concepts or relations randomly, creating the overall set of samples denoted by $\mathcal{T}$. The training objective is a binary classification between positive/negative (denoted as $u$) triplets  with a cross entropy loss function:
\begin{equation}
\begin{aligned}
\mathcal{L}_{\mathcal{G}'} = & -\frac{1}{2|\hat{\mathcal{E}'}|}\sum_{(v_i,r,v_j,u)\in\mathcal{T}}(ulog\ s(v_i,r,v_j) \\
&+(1-u)log(1-s(v_i,r,v_j))).
\end{aligned}
\end{equation}

\subsection{Stance Detection Module}
\textbf{Sentiment Feature Encoding.} To learn sentiment knowledge, we follow \citet{zhou2020sentix} to continually train BERT with sentiment masking. We mask the sentiment-related tokens such as sentiment lexicons, emoticons, and ratings with higher probability than general tokens. The model is trained to reconstruct the masked sentiment tokens and predict the rating of the sentences.  The corrupted text  $\hat{x}$ is fed into BERT to obtain each word representation $\textbf{h}_i$ and the sentence representation $\textbf{h}^{CLS}$. Softmax layers are used on $\textbf{h}_i$ to predict each word's probability, the sentiment of words, and emoticon probability, respectively. A softmax layer on $\textbf{h}^{CLS}$ is also used to predict the rating of the text $\hat{x}$. The tasks are trained using cross-entropy loss. Following \citet{zhou2020sentix} , the SentiBERT are trained on Amazon review dataset \cite{ni2019justifying} and Yelp 2020\footnote{https://www.yelp.com/dataset} challenge dataset.

After pre-training the SentiBERT, given a document $d$ and a topic $t$, we concatenate $d$ and $t$ as our model input $x$ in the following format: $x = [CLS]\ d\ [SEP]\ t\ [SEP]$, SentiBERT to obtain its hidden states:
\begin{equation}
 {\textbf{h}^{fix}_{sent}} = SentiBERT(x),
\end{equation}
where   the parameters of SentiBERT are fixed in our model to keep sentiment information stabilized. 

\textbf{Commonsense Feature Encoding}.
After training the graph autoencoder, in order to extract the document-specific commonsense graph feature for the document $d$ and the topic $t$, the unique nouns, adjectives, and adverbs in the document $d$ and the topic $t$ are extracted at first, which we denote as $S$. Then we extract a sub-graph $\mathcal{G}'_S$ from $\mathcal{G}'$, which contains all the triplets either of whose concepts are in $S$ or within the vicinity of radius 1 from any of the concepts in $S$. Next, we make a forward pass of $\mathcal{G}'_S$ through the encoder of graph autoencoder to obtain the feature vectors $\textbf{h}_j$ for all unique concepts $j$ in $\mathcal{G}'_S$ . The average of  feature vectors $\textbf{h}_j$ for all unique concepts in $\mathcal{G}'_S$ is regarded as the commonsense graph feature vector $\textbf{h}^{KG}$ for the document $d$.
The commonsense graph feature vector $\textbf{h}^{KG}$ is feed into a encoder layer to obtain hidden states $\textbf{h}^K$:
\begin{equation}
    \textbf{h}^K = W_k\textbf{h}^{KG}+b_k
\end{equation}
where $W_k$ and $b_k$ are the trained parameters of the linear layer.

\textbf{Stance Classification.} The input $x$ is first fed into BERT to obtain its hidden states:
\begin{equation}
         {\textbf{h}_{BERT}} = BERT(x)
\end{equation}

Then the hidden states of ${\textbf{h}_{BERT}},  {\textbf{h}^{fix}_{sent}}$ are concatenated and fed into a cross attention module to fuse the information of BERT and SentiBERT:
\begin{equation}
    \textbf{h}^{CLS} = CrossAttention([\textbf{h}_{BERT},\textbf{h}^{fix}_{sent}])[CLS],
\end{equation}
where $\textbf{h}^{CLS}$ is the hidden states of  $[CLS]$ token in BERT.  The hidden states vectors of $\textbf{h}^K$ and  $\textbf{h}^{CLS}$ are concatenated to for classification:
\begin{equation}
    p = Softmax(W[\textbf{h}^{CLS},\textbf{h}^{K}]+b),
\end{equation}
where $W$ and $b$ are the parameters and $p$ is the probability distribution on the three stance labels. 

\textbf{Training.} Given the input and its golden label $(x_i,y_i)$, the loss function $\mathcal{L}_{cls}$ for classifying stance is cross entropy:
\begin{equation}
    \mathcal{L}_{cls} = -\frac{1}{|N|}\sum_{(x_i,y_i)}y_ilog\ p(y_i),
\end{equation}
where $|N|$ is the number of data samples.
To further ensure stronger topic invariance constraints of $\textbf{h}_{KG}$, we add a shared decoder layer $D_{recon}$ with a reconstruction loss:
\begin{equation}
    \mathcal{L}_{recon} = -E_{\textbf{h}^{KG}}(||D_{recon}({\textbf{h}^K}) - \textbf{h}^{KG}||_2^2).
\end{equation}

The overall loss function  is:
\begin{equation}
    \mathcal{L} = \mathcal{L}_{cls}+\mathcal{L}_{recon}.
\end{equation}
   
\section{Experiments}
We verify the effectiveness of sentiment and common sense influence for zero-shot and few-shot stance detection. We also prove the significance of each module in our model in Section 4.4 and analyze the relationship between sentiment (common sense) and stance in Section 4.5 (4.6).
\label{ES}
\subsection{Settings}
\textbf{Dataset}: We adopt the dataset for zero-shot and few-shot stance detection task--VAried Stance Topics (VAST) \cite{allaway2020zero}, which is practical and useful for real-world applications. The dataset consists of thousands of topics, and the statistics are summarized in Table \ref{table1}.  The zero-shot topics only appear in the test set, and the few-shot topics only contain a few training examples. 
\begin{table}[]\small

\begin{center}

\begin{tabular}{cccccccc}
\hline
\hline
      & \#Exp & \#Doc  & \#Zero-shot & \#Few-shot\\
      \hline
Train & 13477                   & 638                & 1481        & 4003               \\
Dev   & 2062                     & 114             & 682         & 383          \\
Test  & 3066                       & 159        & 786         & 600      \\
\hline
\hline
\end{tabular}
\caption{Statistics on the VAST dataset.}
      \label{table1}
\end{center}

\end{table}

\textbf{Training Details} We perform experiments using the official pre-trained BERT model provided by Huggingface\footnote{https://huggingface.co/}. For the pre-trained model with sentiment information, we adopt the model provided by \citet{zhou2020sentix}, which is a continually trained BERT on sentiment datasets. We train our model on 1 GPU (Nvidia GTX2080Ti) using the Adam optimizer \cite{kingma2014adam}. For training the graph autoencoder, the initial learning rate is 1e-2. For the stance detection training process, the initial learning rate is 1.5e-5, the max sequence length for BERT and SentiBERT is 256, the batch size for training is 4, and the model is trained for three epochs.

\begin{table*}[]\small

\begin{center}
 \setlength{\tabcolsep}{2.3mm}{\begin{tabular}{lcccccccccccc}
\hline
\hline
\multirow{2}{*}{Model}            & \multicolumn{4}{c}{ F1 Zero-shot}                                                                                                     & \multicolumn{4}{c}{ F1 Few-Shot}                                                                                                      & \multicolumn{4}{c}{ F1 All}                                                                                                  \\ \cline{2-13} 
                                  & pro                               & con                      & neu                               & all                               & pro                               & con                      & neu                               & all                               & pro                               & con                      & neu                               & \multicolumn{1}{c}{all}  \\ \hline
BiCond                            & .459                              & .475                     & .349                              & .427                              & .454                             & .463                     & .259                              & .392                              & .457                              & .468                     & .306                              & \multicolumn{1}{c}{.410} \\
Cross-Net                         & .462                              & .434                     & .404                              & .434                              & .508                              & .505                     & .410                              & .474                              & .486                              & .471                     & .408                              & \multicolumn{1}{c}{.455} \\
SEKT                              & .504                              & .442                     & .308                              & .418                              & .510                              & .479                     & .215                              & .474                              & .507                              & .462                     & .263                              & .411                     \\
\hline
BERT-sep &.414 &.506&.454&.458&.524&.539&.544&.536&.473&.522&.501&.499\\
BERT-joint                        & .546                              & .584                     & .853                              & .660                              & .543                              & .597                     & .796                              & .646                              & .545                              & .591                     & .823                              & .653                     \\
TGA-Net                           & .554                              & .585                     & .858                              & .666                              & .589                              & .595                     & .805                              & .663                              & .573                              & .590                     & .831                              & .665                     \\
BERT-joint-ft                     & .579                              & .603                     & .875                              & .685                              & .595                              & .621                     & .831                              & .684                              & .588                              & .614                     & .853                              & .684                     \\
TGA-Net-ft                        & .568                              & .598                     & .885                              & .684                              & .628                              & .601                     & .834                              & .687                              & .599                              & .599                     & .859                              & .686                     \\
\hline
Prior-Bin:gold   &\textbf{.643} & .581 &.852  & .692 & .632 &.563 &\textbf{.881} &.692 & \textbf{.652} &.597 &.824 &.691\\
BERT-GCN                          & .583                              & .606                     & .869                              & .686                              & .628                              & .634                     & .830                              & .697                              & .606                              & .620                     & .849                              & .692                     \\
CKE-Net                           & .612                              & .612                     & .880                              & .702                              & \textbf{.644}                              & .622                     & .835                              & .701                              & .629                              & .617                     & .857                              & .701                     \\
\hline
\textit{\textbf{Our Model}} \\
BS                                   & {.625} & {.667}    & {.870}              & {.717}                            & {{.601}}  & {.667}   & {.828}              & {.699}                           & {.591}   & {.669}   & {.858}              &  .706                        \\
S-RGCN                                  & {.582}   & {.669}   & {.838}              & {.699}                           & {.561}  & {.623}    & {.809}              & {.665}                           & {{.607}}    & {.657}  & {.842}              &  .702                        \\
B-RGCN                                      & {.594}  & {.657}    & {.885}              & {.712}                            & {.568}  & \textbf{.678}   & {.851}              & {{.699}}                            & {.591} & {.663}    & {.865}              &      .706                   \\
{BS-RGCN(\textit{proposed})} & {.608}& {\textbf{.674}}  & {\textbf{.895}} & {\textbf{.726}}  & {.600}& {{.665}} & {{.839}} & \textbf{{{.702}}}  & {.604}& \textbf{{.669}} & {\textbf{.866}} & \textbf{.713}   \\
\hline \hline 

\end{tabular}
}
\end{center}
\caption{Overall results. The suffix "ft" means BERT is fine-tuned. BS -- the combination of BERT and SentiBERT; S-RGCN -- the combination of SentiBERT and the graph autoencoder; B-RGCN -- the combination of BERT and the graph autoencoder;  BS-RGCN -- our proposed model. }
\label{results1}
\end{table*}

\textbf{Baselines} We compare our model with several state-of-the-art baselines: (1) \textbf{BiCond} \cite{augenstein-etal-2016-stance}, a model for cross-domain target stance detection task which uses one BiLSTM to encoding the topic and another BiLSTM to encoded the text; (2) \textbf{CrossNet} \cite{xu-etal-2018-cross}, a model based on the BiCond adding an aspect-specific attention layer for cross-target setting; (3) \textbf{SENT} \cite{zhang-etal-2020-enhancing-cross}, a model using the semantic-emotion heterogeneous graph to enhance BiLSTM for cross-traget stance detection;  (4) \textbf{BERT-sep}, a model that encodes the text and topic separately, using BERT, and then classification with a two-layer feed-forward neural network;  (5) \textbf{BERT-joint} \cite{allaway2020zero}, a model with contextual conditional encoding followed by a two-layer feed-forward neural network; (6) \textbf{TGA-Net} \cite{allaway2020zero}, a model using contextual conditional encoding and topic-grouped attention. In addition, we also consider the models BERT-joint-ft and TGA-Net-ft where the BERT module is fine-tuned; (7) \textbf{Prior-Bin:gold} \cite{jayaram021human}, a model applying human rationales as attributions to assist the stance detection; (8) \textbf{BERT-GCN} \cite{liu-etal-2021-enhancing}, a model applying the conventional GCN \cite{kipf2016semi}, which considers node information aggregation; (9) \textbf{CKE-Net} \cite{liu-etal-2021-enhancing}, a model based on BERT, using the CompGCN \cite{vashishth2019composition} to obtain the commonsense information.

\begin{table}[tpb]\small

\begin{center}
\begin{tabular}{cccccc}
\hline
\hline
     Model & Imp & mlT &mlS & Qte & Sarc   \\
      \hline
BERT-joint & .571&.590 &.524&.634&.601                        \\
TGA-Net   &  .594&.605&.532&.661&.637                   \\
BERT-joint-ft  &   .617&.621&.547&.668&.673                  \\
BERT-GCN & .619&.627&.547&.668&.673                \\
CKE-Net & \textbf{.625}&.634&.553&.695&.682               \\
BS-RGCN&.621&\textbf{.647}&\textbf{.556}&\textbf{.701}&\textbf{.717} \\
\hline
\hline
\end{tabular}
\caption{Accuracies on five challenges on the test set. }
\label{breakdown}
\end{center}

\end{table}


\subsection{Results}
The results are shown in Table \ref{results1}. Compared with previous models, our model achieves the state-of-the-art performance in zero-shot, few-shot, and all the topics of VAST. In particular, the macro F1 scores are 72.6\%, 70.2\%, and 71.3\%, which are 2.4\%, 0.1\%, and 1.2\% higher than CKE-Net model, respectively. The results of B-RGCN (our model without SentiBERT module) are 71.2\% and 69.9\%, with a higher macro F1 score on zero-shot topics but a similar result on few-shot topics compared with CKE-Net. The performances of both our model and B-RGCN increase largely on the zero-shot topics but less on few-shot topics, which implies that our graph autoencoder module can achieve a similar effect compared with the GCN module of CKE-Net in the few-shot topics but can improve the effectiveness in extracting relation information in zero-shot topics. This verifies the intuition that only considering the relations between documents and topics limits the transferability of  CKE-Net for the zero-shot task.  Compared with Prior-Bin:gold, the macro F1 scores of our model are 3.4\%, 1.0\%, and 2.2\% higher on zero-shot, few-shot, and all the topics sets, respectively. It implies that commonsense knowledge and sentiment information are more effective than the set of specific human rationales by Prior-Bin:gold as attributions.

 Our model achieves better performance on \textit{Con} labels (67.4\%, 66.5\%, 66.9\%) compared with \textit{Pro} labels (60.8\%, 60.0\%, 60.4\%), which is similar to most of the previous models (BERT-GCN, TGA-Net, and so on).  The phenomenon also appears in B-RGCN and BS, which are our models without SentiBERT and without BERT, respectively (the analysis of the ablation study is explained in Section 4.4 in detail). The results suggest that the use of SentiBERT does not cause the imbalanced performance on different stances and the detection difficulty is mainly on \textit{Pro} labels. In addition, the results of \textit{Neu} stance labels are the highest (89.5\%, 83.9\%, 86.6\%) than those of other labels. It indicates that it is easier for models to classify the \textit{Neu}, where the topics are mostly unrelated to documents.

\subsection{Breakdown Evaluation}
We also test our model on five special phenomena of the test set on VAST following \citet{allaway2020zero}: (1) \textbf{Imp}: non-neutral stance examples where the topics are not explicit in the documents; (2) \textbf{mlT}: documents having multiple stance topics with different topics; (3) \textbf{mlS}: documents having multiple stance topics with different and non-neutral labels; (4) \textbf{Qte}: documents with quotations; (5) \textbf{Sarc}: documents with sarcasm. 

The results are shown in Table \ref{breakdown}. Our model achieves the state-of-the-art performance on mlT, mlS, Qte, and Sarc with 64.7\%, 55.6\%, 70.1\%, and 71.7\%, respectively. In particular, the improvement of our model on mlS implies that different types of knowledge features help models extract stance topics-related information.  The most challenging task is mlS, with a  macro F1 score of 55.6\% by our model. The results demonstrate that it is highly challenging to classify the topics with different stances since the stance information extracted in the model is more related to the whole sentence but more minor to the topics. The macro F1 score of Sarc increases the most, 3.5\% higher than that of CKE-NET, implying that the sentiment information helps boost the model performance in understanding sarcasm, which is a sentiment-related linguistic phenomenon. The accuracy of our model on Imp is the second-highest (slightly lower than that of CKE-Net), which indicates that introducing commonsense graph knowledge can help improve the model performance on the zero-shot task.

      \begin{figure}[tpb]
      \centering
      \includegraphics[width=\hsize]{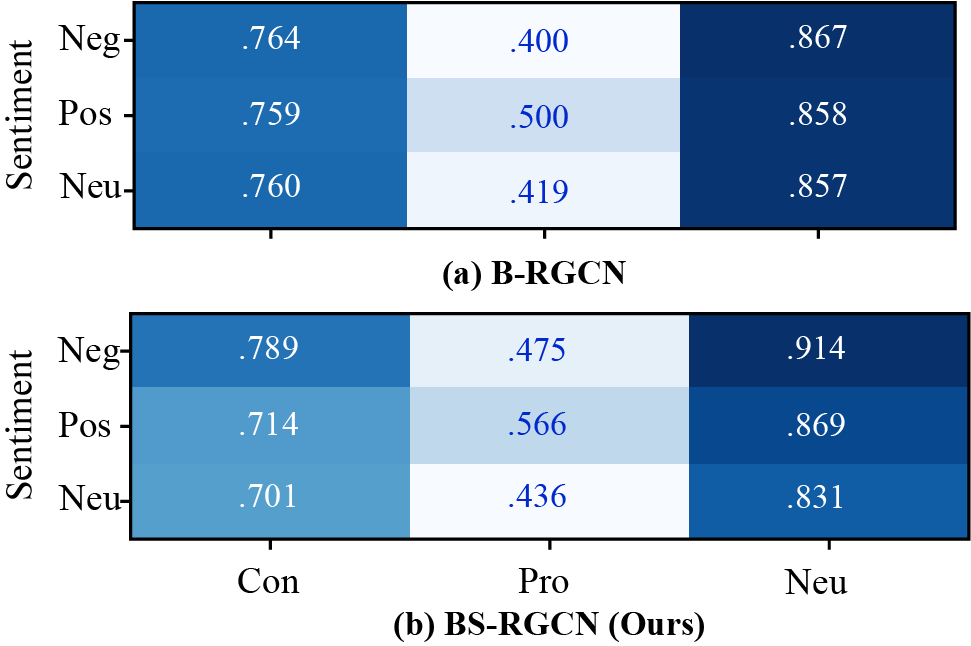}
      \caption{Accuracies of B-RGCN and BS-RGCN on all test data w.r.t different sentiment and stance pairs.}
      \label{fig4}
   \end{figure}

\subsection{Ablation Study}
We conduct ablation studies of BS, S-RGCN, and B-RGCN to understand the significance of the graph autoencoder, BERT, and SentiBERT modules, respectively. The results are shown in Table \ref{results1}. First, BS fuses BERT and SentiBERT feature vectors using Eq(5-6) and classifies the stance using $\textbf{h}^{CLS}$ with a linear layer. It achieves macro F1 scores of 71.7\%, 69.9\%, and 70.6\%  on the zero-shot, few-shot, and all the topics, which are 3.2\%, 1.5\%, and 2.2\% higher than those of BERT-joint-ft, respectively, which proves that sentiment information can help boost the performance of stance detection task. 

Second, B-RGCN and S-RGCN are models without fusing the BERT and SentiBERT feature vectors. The feature vectors of $[CLS]$ tokens from BERT or SentiBERT (the parameters of SentiBERT are not fixed) are directly concatenated with knowledge graph feature vectors to classify the stance. The macro F1 scores of S-RGCN are 69.9\% and 66.5\% on the zero-shot topics and the few-shot topics, 1.3\%, and 3.4\% lower than those of B-RGCN, respectively. It indicates that it is not sufficient to use a sentiment-specific model to do stance classification. The macro F1 score of B-RGCN on the zero-shot set is 71.2\%, 1.0\% higher than that of CKE-Net, which shows that our graph autoencoder module can achieve better performance for zero-shot stance detection than CompGCN. However, BS, B-RGCN, and S-RGCN do not outperform BS-RGCN in the zero-shot topics and all the topics set, which shows that the graph autoencoder, BERT, and SentiBERT are all useful for the stance detection task.

   \begin{table*}[tpb]\small
  \centering
\begin{tabular}{p{0.55\textwidth}p{0.12\textwidth}p{0.13\textwidth}p{0.1\textwidth}}
\hline
\hline
{\textbf{Context}} &\textbf{Topic}& \textbf{Gold Label} & \textbf{Output} \\
\hline

I have lived in brazil for the last five years ( and off and on over the last 27 years ). I know of no one here who is even remotely excited about the Olympics. It would seem that people don\'t care. The economy is {\color{teal}tanking} and government is at a complete {\color{teal}standstill}. We have more important things on our mind right now. &  Olympics &  Con &Con \\ 
\hline
I can't even believe that this is a debate. Cutting the most basic \textbf{foreign language programs}? How does one appreciate that there is a world outside of America? Google translate? Suny, everyone is laughing at you and you're too smug to notice. &College&Con&Con \\ 
\hline
Good idea. I have always had a cat or two. While being \textcolor{teal}{inhumane}, \textbf{declawing} places a cat in danger. Should my charming indoor kitty somehow escape outside, he would have no way to defend himself. 
& nail removal &Con &Con\\

\hline
\hline
\end{tabular}
\caption{Case Study for our trained stance detection model. Case I shows the effectiveness of using sentiment information; Case II shows the importance of commonsense knowledge; Case III shows both the sentiment and commonsense knowledge help the stance detection model.}
\label{case}
\end{table*}

\subsection{Sentiment and Stance}

\citet{allaway2020zero} indicate that models of BERT-Joint are reliant on sentiment cues, and the models  learn the strong association between the \textit{Neg} (negative) sentiment and the \textit{Con} stance, yet weak association between \textit{Pos} (positive) sentiment and \textit{Pro} stance. Their analysis is based on experiments where the documents are corrupted by replacing the text's sentiment words.
Here we take a different perspective and carry out experiments with respect to different stances and sentiment pairs on both B-RGCN and BS-RGCN. We use opinion lexicon \cite{hu2004mining} to classify the sentiment of document, (i.e, if a document contains more positive/negative words, we treat it as a  document with the \textit{Pos} (positive)/\textit{Neg} (negative) sentiment; otherwise, we treat it as a  document with the \textit{Neu} (neutral) sentiment).

The results are shown in Figure \ref{fig4} (the model trained on all the topics is tested in this experiment). For BS-RGCN, the accuracy on the corresponding stance and sentiment \textit{(Neg, Con)} is 78.9\%, higher than 71.4\% of  \textit{(Pos, Con)}  and 70.1\% of \textit{(Neu, Con)}. Similarly, the accuracy on \textit{(Pos, Pro)} is 56.6\%, higher than 47.5\%   of \textit{(Neg, Pro)} and 43.6\% of \textit{(Neu, Pro)}. This suggests that data samples with corresponding sentiment and stance pairs (\textit{(Pos, Pro), (Neg, Con)}) are easier to classify by our model. The performance of B-RGCN is similar to BS-RGCN, with an accuracy of 76.4\%  for  \textit{(Neg, Pro)}, a little higher than those of \textit{(Pos, Con)} (75.9\%) and \textit{(Neu, Con)} (76.0\%). The same model  achieves an accuracy of 50\%  of \textit{(Pos Pro)}, 10\% higher than that of \textit{(Neg, Pro)}, and 8.1\% higher than that of \textit{(Neu, Pro)}. The model without the sentiment module can also predict corresponding sentiment and stance pairs with higher accuracy, demonstrating that sentiment information can help stance detection models. The accuracies for B-RGCN and BS-RGCN are both significantly higher on data with \textit{Con} stances than those with \textit{Pro} stances. The phenomenon indicates that it is difficult for models to predict \textit{Pro} stance in the VAST dataset, and the difference in performance is not caused by the difference of associations between data of \textit{(Pos, Pro)} and \textit{(Neg, Con)}.  For the data of \textit{Neu} stance, the performance is less related to sentiments. The models can achieve much better results on \textit{Neu} stance data, where the topics may be not related to the documents, 91.4\% on \textit{(Neg, Neu)}, 86.9\% on  \textit{(Pos, Neu)}, 83.1\%  on \textit{(Neu, Neu)} for BS-RGCN , and 86.7\%  on \textit{(Neg, Neu)}, 85.8\% on  \textit{(Pos, Neu)}, 85.7\%  on \textit{(Neu, Neu)} for B-RGCN. The phenomenon demonstrates that it is easy for the model to judge whether the topic is related to the documents.
      \begin{figure}[tpb]
      \centering
      \includegraphics[width=\hsize]{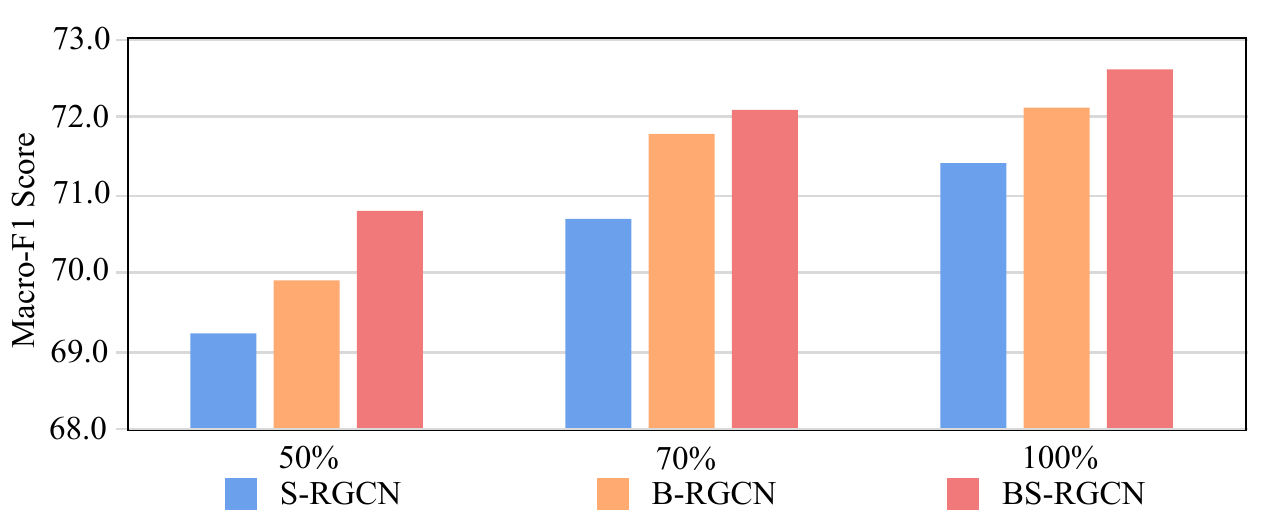}
      \caption{Macro-F1 scores of S-RGCN, B-RGCN, and BS-RGCN on zero-shot test data w.r.t different percents of commonsense knowledge for pre-training.}
      \label{commonsense}
   \end{figure}

\subsection{Common Sense and Stance}
We show the relationship between common sense and stance by pre-training the graph autoencoder w.r.t different percentages of extracted concepts (Section 3.1). Using the commonsense feature with the pre-trained autoencoder, we show the results of the stance detection models B-GCN, S-GCN, and BS-RGCN on the zero-shot task. The results are given in Figure \ref{commonsense}. As observed, the performance of the three models increases with increasing coverage of commonsense knowledge. It indicates that commonsense knowledge is directly useful for stance detection models.

\subsection{Case Study}
We also show some cases from the test data using the model trained on all the topics. In the first case, sentiment words such as  `tanking' or `standstill' imply the negative sentiments towards the influence of the Olympics on the economy of Brazil, which further expresses an opposing stance towards `Olympics'. Our model outputs the correct label towards the target thanks to the sentiment information. In the second case, no explicit expression of the target `College' is contained in the document. Only some implications, including `foreign language programs', have relation to the `College', and with the commonsense knowledge encoding, our model outputs the correct stance. The third case proves that both common sense and sentiment information can benefit the stance detection model, that `inhumane' expresses a negative sentiment, and the topic `nail removal' is implicitly 
involved by the word `declawing'.  Our model can also give the correct stance for case III. 



\section{Conclusion}
We proposed a stance detection model incorporating commonsense knowledge and sentiment information, achieving state-of-the-art zero-shot and few-shot stance detection results on the standard dataset. The ablation study showed the significance of each module, such as knowledge graph autoencoder, SentiBERT, and BERT.  We also analyzed the relation between sentiment/common sense and stance, which indicate the effectiveness of this external knowledge.


\section*{Acknowledgements}
Yue Zhang is the corresponding author. We would also like to thank the anonymous reviewers for the detailed and thoughtful reviews. The work is funded by the Zhejiang Province Key Project 2022SDXHDX0003.

\bibliography{anthology,custom}

\appendix
\section{Appendix}
\textbf{Human Labeling for Sentiment and Stance Detection}

In this part, we manually label some samples (randomly selected) from VAST dataset to prove the relation between sentiments and stances. Opinion lexicon \cite{hu2004mining} is adopted as the sentiment vocabulary. As shown in Table \ref{case2}, there are many samples (7 in 10) that sentiment knowledge plays a significant role for stance detection, and few samples have a conflicting relation.

 \begin{table*}[h!]\small
  \centering
\begin{tabular}{p{0.65\textwidth}p{0.1\textwidth}p{0.05\textwidth}p{0.1\textwidth}}
\hline
\hline
{\textbf{Context}} &\textbf{Topic}& \textbf{Stance} & \textbf{Relation} \\
 
\hline 
The reason that \textbf{Deep Mind} winning is so \textcolor{red}{impressive} is that Google managed to accomplish this 
with virtually no warning. It was less than a year ago where the \textcolor{red}{best}\textbf{ computer program} was not in the 
top 20,000 in the world. It was less than 6 months ago when the program beat a player in the top 1,000.  Yesterday the program beat the the best player in the world. Am I wrong to be shocked at how \textcolor{red}{fast} complicate \textbf{AI} has advanced? & Artificial Intelligence  &  Pro &  + \\

\hline
I totally agree with this premise. As a younger person I was against \textbf{Nuclear power} (I was in 
college during 3 mile island) but now it seems that nuclear should be in the mix. Fission technology 
is \textcolor{red}{better}, and will continue to get \textcolor{red}{better} if we actively promote its development. The \textcolor{red}{prospect} of fusion
energy also needs to be explored. If it's \textcolor{red}{good} enough for the sun and the stars, it's \textcolor{red}{good} enough for me. &Nuclear Power & Pro & + \\
\hline
This is a horrible idea. Anyone who has worked on the border, or in \textbf{Mexico} (as I do), knows there are plenty of middle and upper-class Mexicans who come to the U.S. for an education. I think Dr. Lee is really perpetuating stereotypes here. In my opinion, affirmative action should be based on economic class, no matter what the race. & Mexico & Pro &0 \\
\hline
Good idea. I have always had a cat or two. While being \textcolor{teal}{inhumane}, \textbf{declawing} places a cat in danger. Should my charming indoor kitty somehow escape outside, he would have no way to defend himself. Why don't humans have their finger-and tonails removed to save on manicures? Answer: they are important to the functioning and protection of our bodies. & nail removal & Con & +\\
\hline
The mandate of \textbf{private corporations }is to make a profit. And if the profit is made at the EXPENSE of the society that allow the \textbf{corporation} to exist, well,  \textcolor{teal}{too bad}. \textbf{Oil companies} \textcolor{teal}{foul} the environment. \textbf{Financial companies} drive the economy into the Great \textcolor{teal}{Recession}. \textbf{Airlines} have no regard for the people they transport. As long as they make a profit, they are allowed to \textcolor{teal}{abuse} the public until they are stopped. That is the way it has been since Swift and Armour canned and sold rotten meat and Carnegie sent Pinkertons to shoot striking miners.  & private corporation profit &Con &+ \\
\hline
One's own, and learning another language is important and a great work out for the brain! Back in the day, I learned Spanish! In retrospect \textbf{Latin} would have been the \textcolor{red}{better} way to go, since mastery of that makes learning the languages like French, Italian, Portuguese, Romanian, and Spanish, very much easier to learn!  &Latin helpful language &Pro &+\\
\hline
Without government to ensure their behavior, \textbf{companies} will attempt to make a profit even to the DETRIMENT of the society that supports the business. We have seen this in the environment, in finances, in their treatment of workers and customers. Enough. &company &Pro&0\\
\hline

The "you have a short live, so enjoy" attitude alone did not lead to the \textbf{Renaissance}, the age of Enlightment, or the Industrial Revolution. It did not le ()ad to the invention of the light bulb, or the telephone, or the internet, or the NYT electronic discussion board. Just "enjoying" life alone means you are enjoying the fruit of someone else\'s hard work. &Renaissance &Pro&0
\\
\hline
Of course their salaries should be raised. But this should be separated from the discussion about legality. Salaries should be raised and only legal workers should be employed. Its really a no brainer. And any discussion about only Mexicans being prepared to do this work so it has to be \textbf{illegal} is completely \textcolor{teal}{disingenuous}. & illegal labor &Con &+ \\
\hline 
Also, and usually not acknowledged, is that we are slowly eroding the fertility of the soil. There is no more usable \textbf{soil}, we are farming everything that can be farmed. Current methods depend on petrochemical fertilizers. Even with their use, \textbf{fertility} is slowly \textcolor{teal}{declining}. As human population continues to grow, the result is obvious. &soil &Con&+ \\
\hline
\hline
\end{tabular}
\caption{Manually labeling samples for the relation between sentiments and stances. The positive/negative words related to the topics are labeled with red/teal colors. The topic related words in the documents are bold. `+' for sentiment words supporting the stance, `0' for no relation.}
\label{case2}
\end{table*}

\end{document}